# Probabilistic Models for Anomaly Detection in Remote Sensor Data Streams

**Ethan W. Dereszynski**
Computer Science Dept.
Oregon State University
Corvallis, OR 97330

**Thomas G. Dietterich**
Computer Science Dept.
Oregon State University
Corvallis, OR 97330

## Abstract

Remote sensors are becoming the standard for observing and recording ecological data in the field. Such sensors can record data at fine temporal resolutions, and they can operate under extreme conditions prohibitive to human access. Unfortunately, sensor data streams exhibit many kinds of errors ranging from corrupt communications to partial or total sensor failures. This means that the raw data stream must be cleaned before it can be used by domain scientists. In our application environment—the H.J. Andrews Experimental Forest—this data cleaning is performed manually. This paper introduces a Dynamic Bayesian Network model for analyzing sensor observations and distinguishing sensor failures from valid data for the case of air temperature measured at 15 minute time resolution. The model combines an accurate distribution of long-term and short-term temperature variations with a single generalized fault model. Experiments with historical data show that the precision and recall of the method is comparable to that of the domain expert. The system is currently being deployed to perform real-time automated data cleaning.

## 1 INTRODUCTION

The ecosystem sciences are on the brink of a huge transformation in the quantity of sensor data that is being collected and made available via the web. Old sensor technologies that measure temperature, wind, precipitation, and stream flow at a small number of spatially distributed stations are being augmented by dense wireless sensor networks that can measure everything from sapflow to gas concentrations. Data streams from existing and new sensor networks are being posted directly to the web. The resulting explosion in data is likely to transform ecology from an analytical and computational science into a data exploration science (Gray & Szalay, 2002).

Unfortunately, raw sensor data streams can contain many kinds of errors. Sensors can be damaged by extreme weather, information can be corrupted during data transmission, and environmental conditions and technical errors can change the meaning of the sensor data (e.g., an air temperature sensor buried in snow is no longer measuring air temperature, two thermometers whose cables are swapped during maintenance will not be measuring the intended temperatures, etc.). In current practice, data streams undergo a quality assurance (QA) process before they are made available to scientists. This is typically a manual process in which an expert technician visualizes the data in various ways looking for outliers, unusual events, and so on. But this manual approach has two obvious drawbacks. First, it is slow, expensive, and tedious. This introduces a substantial delay (3-6 months) between the time the data is collected and the time the data is made publicly available. Second, it will not scale up to the large amounts of data that will be collected by dense sensor networks. Hence, there is a need for automated methods for "cleaning" the data streams to flag suspicious data points and either call them to the attention of the technician or automatically remove incorrect values and impute corrected values.

This paper describes a Dynamic Bayesian Network (DBN, Dean & Kanazawa, 1988) approach to automatic data cleaning for individual air temperature data streams. The DBN combines discrete and conditional linear-Gaussian random variables to model the air temperature at 15 minute intervals as a function of diurnal, seasonal, and local trend effects. Because the set of potential faults is unbounded, it is not practical to approach this as a diagnosis problem where each fault is modeled separately. Instead, we employ a very general fault model and focus our efforts on making the DBN model of normal behavior highly accurate. The hope is that if the observed temperature is unlikely based on the temperature model, the fault model will become more likely. The DBN contains two hidden variables: the current state of the sensor and the current temperature trend (as a departure from the baseline temperature). The model is applied online as a filter to decide the state of the sensor at each 15 minute point. If the sensor is believed to be bad, the observed temperature is ignored by the DBN until the sensor returns to normal. As a side effect, the model predicts what the true temperature was during periods of time when the sensor is bad.

Belief Networks (Pearl, 1988) have been employed for sensor validation and fault detection in domains such as robotic movement, chemical reactor, and power plant monitoring (Nicholson & Brady, 1992; Mehranbod et al., 2003; Ibarguengoytia et al., 1996). Typically, the uncertainty in these domains is limited to the sensor's functionality. That is, the processes in these domains function within some set boundaries with a behavior that can be modeled by a system of known equations (Aradhye, 1997). Ecological domains are challenging, because accurate process models encompassing all relevant factors are typically unavailable (Hill & Minsker, 2006). Thus, uncertainty must be incorporated into both the process and sensor models.

This paper is organized as follows. We first discuss the nature of the temperature data, the sensor sites at the H.J. Andrews Experimental Forest, and the anomaly types encountered. Second, we describe the temperature prediction model, including training and inference. Finally, we present the results of the model applied to temperature data from the Andrews.

## 2 APPLICATION DOMAIN

### 2.1 ATMOSPHERIC DATA AT THE ANDREWS

We focus our application on twelve air temperature sensors distributed over three meteorological stations at the H.J. Andrews Experimental Forest, a Long Term Ecological (LTER) site located in the central Cascade Region of Oregon. The three meteorological stations—Primary, Central, and Upper Lookout—are located at elevations of 430 meters, 1005 meters, and 1280 meters, respectively. Each site contains four air temperature sensors mounted on a sensor tower. The sensors are placed at heights of 1.5 meters, 2.5 meters, 3.5 meters, and 4.5 meters above ground level. The sensors record the ambient air temperature every 15 minutes and transmit the recorded value to a data logger located at the meteorological ("met") station. The logger periodically transmits the batch data back to a receiving station at the Andrews Headquarters, where it is reviewed by a domain expert before being made available for public download (McKee, 2005). There are 96 observations (quarter-hour intervals) per day and 35,040 observations per year. During the winter and during strong storms and floods, the Central and Upper Lookout stations are usually inaccessible.

The observed air temperature data contains significant diurnal (time of day) and seasonal (day of year) effects. Temperature rises in the morning and falls in the evening (the diurnal effect). Temperatures are higher in the summer and colder in the winter (the seasonal effect). These effects interact so that in the summer, the diurnal effect is more pronounced—the temperature swings are larger and the temperature rises and falls faster—than in the winter. In addition, weather systems (cold fronts, heat waves) cause medium term (1-10 day) departures from the temperatures that would be expected based only on the diurnal and seasonal effects. Figure 1 illustrates diurnal, seasonal, and weather effects on air temperature.

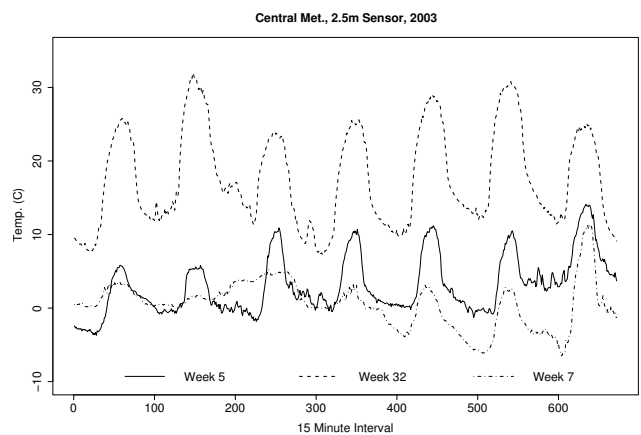

Figure 1: Seasonal, Diurnal, and Weather effects

Week five and week thirty-two demonstrate the seasonal effect on air temperature (winter and summer, respectively), whereas week seven illustrates a weather event suppressing the diurnal effect for the first four days of the week.

### 2.2 DEGREES OF ANOMALY

We classify anomaly types found in the air temperature data into three categories based on subtlety of the anomaly in the context of the data.

#### 2.2.1 Simple Anomalies

We consider simple anomalies to be observations far outside the range of acceptable temperatures. These anomalies are introduced deliberately by the data logger. If the sensor is disconnected from the data logger, the logger records a value of $-53.3$. If the logger receives a voltage outside the measurement range for the sensor, the logger records a value of $-6999$. These two are the most common anomaly types, because sensor disconnections and damage to the wiring may persist for long periods of time depending on the accessibility of the sensor. Figure 2 contains two such examples.

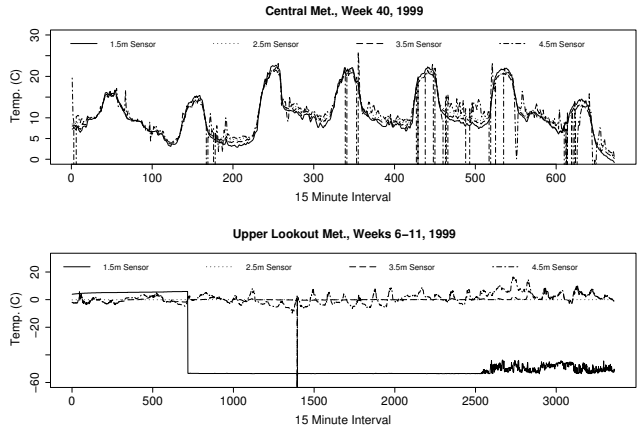

Figure 2: Top: 4.5m voltage out of range, $-6999.9$ fault value. Bottom: 1.5m disconnected from logger, $-53.3$ fault value

### 2.2.2   Medium Anomalies

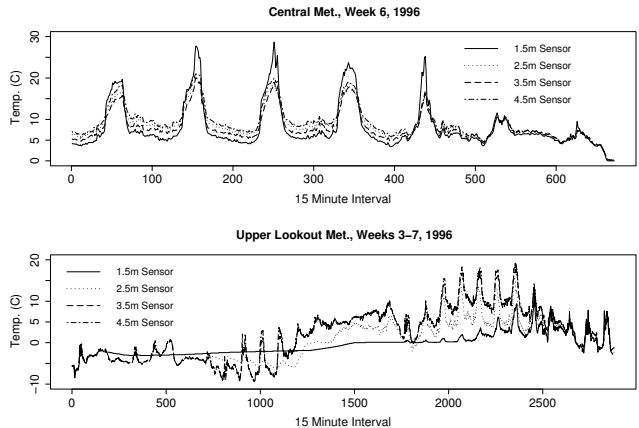

Figure 3: Top: Broken Sun Shield, Bottom: 1.5m Sensor buried under snowpack, 2.5m Sensor dampened

This anomaly type is associated with malfunctions in the sensor hardware or change in functionality of the sensor. For example, if the sensor's sun shield becomes damaged or lost, then direct sunlight exposure introduces a positive bias in the recorded value. These anomaly types are correlated with external weather conditionsand hence contain many of the same trend effects as the valid data, which makes them harder to detect. Figure 3 contains two such examples.

The first plot illustrates the loss of a sun shield on the 1.5m sensor, which raises the recorded temperature by approximately 5 ℃. It is important to note that this bias disappears during the nighttime periods and also on cloudy days (which probably explains why the bias is missing during the last two days of the week).

The second plot results from a snowpack that has buried the 1.5m sensor by the 200th quarter-hour measurement. This sensor records the temperature as a near-constant $-2$ ℃ for approximately 3 weeks. Notice that the 2.5m sensor is also affected by the snow: its diurnal behavior is significantly dampened. Indeed, we can observe that the snow first buries the 1.5m sensor before affecting the 2.5m sensor and that, as the snowpack melts, the 2.5m sensor returns to nominal behavior before the 1.5m sensor.

### 2.2.3   Complex Anomalies

We reserve this classification for anomalies that are so subtle that they cannot be captured without the use of additional sensors. An example of a complex anomaly is a switch in sensor cables between two adjacent sensors on a tower. Because under normal conditions the two sensor readings differ by only a fraction of a degree Celsius, if we examine only one of the sensor streams, we cannot detect the anomaly. However, a model of the joint distribution of all four sensors on the tower should be able to capture the fact that the relative order of the sensor values reflects their physical order on the tower. Specifically, the 4.5m sensor is the hottest of the four in the mid-afternoon and the coolest of the four in the middle of the night. Because the present paper only models individual sensor streams, we do not expect it to detect these complex anomalies.

## 3   DOMAIN MODELING

We employ a Conditional Linear Gaussian (CLG) (Lauritzen, 1992) DBN to model the interaction between a single sensor and the air temperature process. Each *slice* of temporal model represents a fifteen-minute interval, as this is the time granularity of our sensor observations. However, our model also contains two observed discrete variables representing the current time window:

1. $QH = 1, ..., 96$ Representing the current quarter-hour of the day

2. $Day = 1, ..., 365$ Representing the current day of the year.

Thus, every *slice* is associated with a unique $(QH = qh, Day = d)$ pair. Further discussion will be split into the process model governing air temperature change and the model for sensor behavior.

## 3.1  THE PROCESS MODEL

For any given time step and $(qh, d)$, we assume the actual temperature, $T$, is a function of some *learned* baseline value, $B$ (please see the below section for discussion of the baseline value), and a value representing the current departure from the baseline value, $\Delta$. That is, we estimate the distribution over $T$ as $T \sim N\left(\Delta + B, \sigma_T^2\right)$.

The $\Delta$ variable can be interpreted as representing a temporally local trend effect, such a warm/cold front or a storm. Its purpose is to capture the difference between our baseline expectation for the temperature at a given time of day/day of year and the observed temperature during periods of nominal sensor behavior. We model $\Delta$ as a first-order Markov process with the current $(qh, d)$ as additional non-Markovian, observed inputs. Thus, $\Delta$ has the distribution $\Delta \sim N\left(\mu_{qh,d} + w\Delta_{t-1}, \sigma_{qh,d}^2\right)$. The Markov process allows the $\Delta$ distribution to "wander" in order to capture growing or diminishing trend effects. By conditioning the distribution on $QH$ and $D$, we can account for sharper temperature shifts associated with particular $(qh, d)$ pairs. For example: the temperature rises and falls more quickly as a result of diurnal effects in summer months than in winter months. To account for this, $\Delta$ must be able to change more rapidly (have increased variance) during these periods of the day and season.

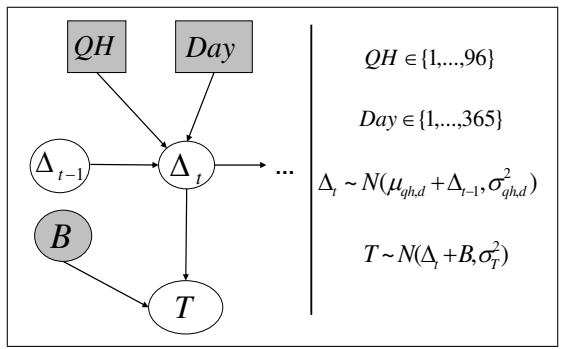

Figure 4: Process model for air temperature. Rectangles depict discrete variables, ovals depict normally distributed variables. Shaded variables are observed.

### 3.1.1  Calculating the Baseline

The baseline value for a particular $(qh, d)$ pair estimates the temperature for that time interval after removing short-term trends due to weather systems. Initially, it may seem appropriate to simply average temperature values for a given $(qh, d)$ pair across all of the training years; however, as we have only a few training years, there is too much variance in sample means to provide a good estimate. To address this, we apply a moving average kernel smoother across the $M$ days on either side of the current day and the $N$ quarter-hour periods on either side of the current quarter hour. However, if we only used this simple smoother, it would be biased low at times when the second derivative of the temperature was negative (at the point of maximum temperature) and biased high at times when the second derivative of the temperature was positive. To correct for this, we compute the first derivative $Q(d, qh, t, y)$ for each $(u, t)$ offset, and use this to remove the short-term linear trend in the temperature curve:

$$B_{qh,d} = c \sum_{y,u,t} T\left(d + u, qh + t, y\right) - Q\left(d, qh, t, y\right)$$

$$c = \left[Y\left(2M + 1\right)\left(2N + 1\right)\right]^{-1},$$

$y \in \{1, ..., Y\}$ denotes the year index.
$u \in \{-M, ..., M\}$ denotes the day offset.
$t \in \{-N, ..., N\}$ denotes the quarter-hour offset.
$T(d, qh, y)$ training value for given $(d, qh, y)$ tuple.

$Q(d, qh, t, y)$ is the first-derivative offset function that calculates the average deviation from the current quarter-hour to $t$ over a $2M + 1$ day window. It is calculated as $Q(d, qh, t, y) =$

$$(2M + 1)^{-1} \sum_u T\left(d + u, qh + t, y\right) - T\left(d + u, qh, y\right).$$

## 3.2  INTEGRATING THE SENSOR

Based on our discussions with the expert, we have identified several anomaly types in the temperature data streams. However, we are not confident that we have found all anomaly types. Each time we meet with the expert, we learn about a new anomaly, and there is no reason to expect that the set of anomalies is fixed. Hence, rather than attempting to model each type of anomaly separately, which would lead to a system that could only recognize a fixed set of anomaly types, we decided to develop a single, very general fault model that is able to capture most of the known anomaly types and (we hope) unknown types as well. We model the state of the sensor, $S$, as a discrete variable that summarizes the degree of sensor functionality. We chose four levels ("very good", "good", "bad", "very bad"), because these are the levels that the expert technician uses in the current QA process. The state $S$ is modeled as a first-order Markov process, which allows us to capture the fact that good sensors tend to stay good and bad sensors tend to stay bad. The observed temperature, $O$, is distributed as $N(\mu_s + w_s T, \sigma_s^2)$, where the values of $\sigma_s^2$ capture how well the observed temperature is tracking the true temperature as a function of the current state $S = s$. Figure 5 depicts the full domain model.

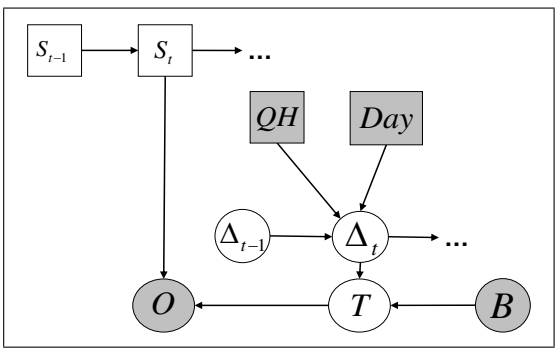

Figure 5: Integrated single-sensor and process model.

# 4    METHODS

We obtained seven years of raw air temperature from the H. J. Andrews Experimental Forest. This data had been processed by the domain expert to mark all anomalous data points. The domain expert tends to mark regions as anomalous even when some of the readings within those regions are ok. For example, if a sun shield was missing, the expert would mark the entire time interval when the shield was missing as anomalous even though at night time and on cloudy days the temperature readings are accurate.

## 4.1    PARAMETERIZATION

For each of the meteorological stations, we selected three years of data as our testing set. From this set we removed all data points labeled anomalous by the expert, and trained on the remaining data. We calculated a baseline value for every $(qh, d)$ pair as described in section 3.1.1, with the number of days $M = 3$ and the number of quarter hours $N = 5$. Using these values, we then iterated through the training set and calculated $\Delta(y, d, qh)$ as $T(y, d, qh) - B(d, qh)$, where $T(y, d, qh)$ is the recorded temperature for that year, day, and quarter-hour, and $B(d, qh)$ is the baseline.

To fit the conditional distribution for $\Delta$, we smooth over a 31-day window. We compute the mean of $\Delta_{qh,d}$ as

$$\mu_{qh,d} = (YM)^{-1} \sum_{y,u} \Delta(y, d+u, qh) - \Delta(y, d+u, qh-1)$$

and the variance as

$$\sigma^2_{qh,d} = (YM)^{-1} \sum_{y,u} \left( \Delta(y, d+u, qh) - \mu_{qh,d} \right)^2,$$

where $Y = 4$ is the number of years, and $M = 31$ is the number of days.

We manually tuned the parameters for the predicted temperature $(T)$, the observed temperature $(O)$, and the sensor $(S)$ variables to implement the generic fault model. The sensor can be in one of four states: {Very Good, Good, Bad, Very Bad}. The first three states assert equality between the mean of the predicted and observed temperatures, and the last state encompasses anomalies with no correlation to $T$.

Table 1: Observed Temperature as a function of Sensor State and Predicted Temperature

| SENSOR STATE | DISTRIBUTION |
|---|---|
| $O\|S_t = VeryGood$ | $N(T, 1.0)$ |
| $O\|S_t = Good$ | $N(T, 5.0)$ |
| $O\|S_t = Bad$ | $N(T, 10.0)$ |
| $O\|S_t = VeryBad$ | $N(0, 100000)$ |

We calculate the actual temperature as the sum of the baseline and the current $\Delta$ offset. We assign weights of 1.0 for both these variables and set $\Delta$'s mean to 0. Further, we supply $\Delta$ with a very-low, non-zero variance. The distribution over the observed temperature is tied to the state of the sensor, and thus we can use the sensor state to explain large residuals between the observed and predicted temperature. In cases where the sensor is believed to be functioning nominally, the observed temperature should be the predicted temperature with some minimal variance. If we believe the sensor is malfunctioning, we allow the observed temperature to take on additional variance yet still reflect the mean of the predicted temperature. In cases where we believe the sensor is completely failing, we set the weight from the predicted temperature to 0, and assign a huge variance to $O$. Table 1 displays the distribution of $O$ given the state of $S$.

## 4.2    INFERENCE

We perform inference in our network using a variation on the Forward Algorithm (Rabiner, 1989) and Bucket Elimination adapted for CLG networks (Dechter, 1996; Lauritzen, 1992). The Forward Algorithm computes the marginal for every step of a Markov process and passes that distribution forward as the *alpha message*. The Bucket Elimination algorithm is a dynamic-programming, exact-inference method that represents potentials as buckets and marginalizes out variables iteratively until only the desired potential remains.

Table 2 outlines our modified Forward inference method. The two modifications to the Forward algorithm occur in steps 3 and 5. In 3, we enforce a decision about the state of the sensor and use its most likely value to constrain the distribution on $\Delta$ computed in 4. In other words, at each step, we compute the posterior distribution over $S$ and then force $S$ to take on its most likely value. This is necessary to prevent the variance associated with $\Delta$ from growing rapidly.

Table 2: Modified Forward Algorithm

1. Enter evidence for observed variables: $QH$, $Day$, $B$, and $O$

2. Compute the posterior for $S_t$ as the new alpha message, $\alpha_S$

3. Enter the most likely value of $S_t$, $\overset{argmax}{s} P(S_t = s | O_{1:t})$, as the data label for time $t$ and as additional evidence.

4. Compute posterior for $\Delta_t$ as the new alpha message, $\alpha_\Delta$

5. If $s = $ Very Bad, then set variance of $\alpha_\Delta$ to $\min\left(\sigma_{qh,d}^2, \sigma_x^2\right)$ where $\sigma_{qh,d}^2$ is the regular variance of $\Delta$ for that $(qh,d)$ pair and $\sigma_x^2$ is the calculated variance of $\alpha_\Delta$

6. Update $S_{t-1}$ to $\alpha_Q$ and $\Delta_{t-1}$ to $\alpha_\Delta$ (pass $\alpha$ messages forward) and return to 1.

Consider the elimination ordering in Table 3. The potential in (1) will fail to sufficiently constrain the variance of $\Delta_t$, because there is always a non-zero probability that $S_t = $ Very Bad. The high variance associated with the general fault state of the sensor then removes the constraining effect the observation $O$ provides. By entering evidence for the sensor state before computing the posterior of $\Delta_t$, we eliminate this problem. The additional observation for $S_t$ changes the expression in (1) to:

$$P(O = o | S_t = s, T)P(T | B = b, \Delta_t) = Pot(\Delta_t)$$

and removes the potential calculated in (4). $Pot(\Delta_t)$ sufficiently constrains $\Delta_t$'s variance for all values of $s_t$ except *Very Bad*. We address the latter case in step 5 of our algorithm (Table 2) by setting an upper limit on the variance of the posterior. The limit is the trained variance parameter for $\Delta$ for the current $(qh,d)$ pair.

Table 3: Bucket Elimination for $P(\Delta_t)$ Elimination Ordering: $T, \Delta_t, S_t, S$

$$P(O = o | S_t, T) P(T | B = b, \Delta_t) = Pot(\Delta_t | S_t) \quad (1)$$
$$P(\Delta_t | \Delta_{t-1}) P(\Delta_{t-1}) = Pot(\Delta_t) \quad (2)$$
$$P(S_t | S_{t-1}) P(S_{t-1}) = Pot(S_t) \quad (3)$$
$$Pot(\Delta_t | S_t) Pot(S_t) = Pot(\Delta_t)' \quad (4)$$
$$Pot(\Delta_t) Pot(\Delta_t)' \quad (5)$$

## 5   RESULTS AND ANALYSIS

We test our method over four years of labeled data from the H.J. Andrews (3 held out for training). Training and testing years were individually selected for each site with a preference for years including few or no anomalies in the training set, and years exhibiting the largest diversity of anomaly types used for testing. We analyze our results in terms of anomaly difficulty, and then introduce an additional classification method designed to approximate the behavior of the domain expert. Finally, we report overall precision and recall with regard to our classification system.

### 5.1   SIMPLE ANOMALIES

Figure 6 shows a typical result provided by the model for intermittent sensor faults associated with a voltage error in the sensor. We omit the *Good* and *Very Good* labels for clarity. Note that all values of $-6999.9$ ℃ were correctly labeled as *Very Bad*; also, the model produced no false positives for the week shown. The predicted temperatures values inserted in place of those labeled anomalous closely resemble the neighboring valid segments of the data stream.

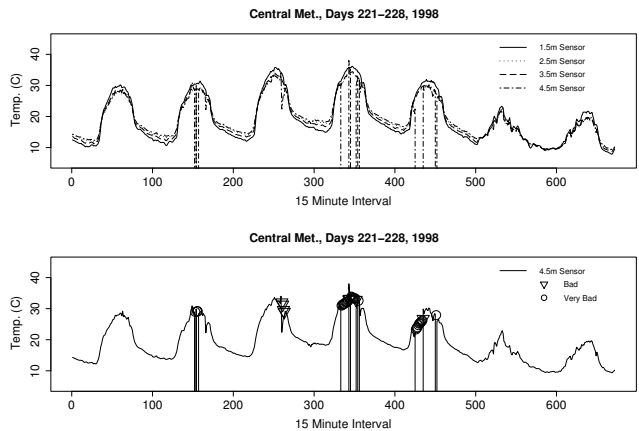

Figure 6: Top: Original data stream containing a faulty 4.5m sensor. Bottom: Data cleaning results for the 4.5m sensor. Plotted points indicate the mean of the predicted temperature distribution.

### 5.2   MEDIUM ANOMALIES

Let us consider the broken sun shield anomaly introduced in Section 2.2.2. Figure 7 (bottom) shows the labeling provided by our method on an anomalous section of data. The model correctly identifies the ascending and descending segments of the day as anomalous, while labeling those periods unaffected by the broken sun shield (nighttime and periods of cloud cover) as normal. On some days, the model correctly labels the peak of the diurnal period as anomalous, but in other cases (e.g., $t$=260), it does not. This is because the short-term behavior at the peak looks normal (except for its absolute value): the temperature reaches its high for the day, it holds steady for a short period, and then begins to decrease. The reduced rate of change in temperature between time slices then falls within the range of $\Delta$'s variance, so it is labeled as non-anomalous. Note that the model-predicted temperature slightly lags the observed 4.5m temperature. This matches the 1.5m, 2.5m, and 3.5m sensors, which are labeled by the domain expert as functioning nominally for this period and corrects for the incorrect acceleration/deceleration introduced by the fault.

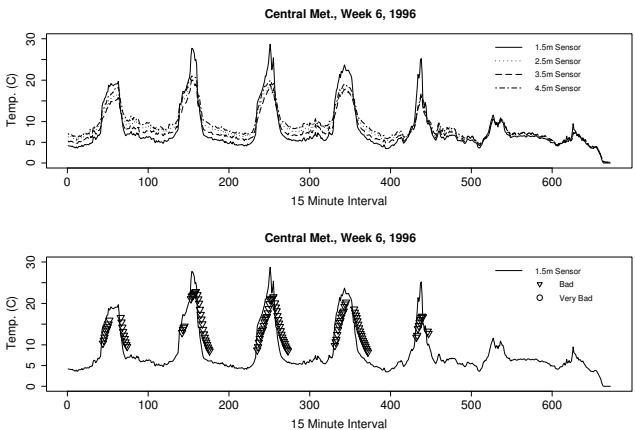

Figure 7: Top: Lost sun shield in 1.5m sensor. Bottom: Data cleaning applied to 1.5m sensor. Points indicate the mean of the predicted distribution.

## 5.3  CLASS WIDENING

As a result of working at very fine time granularities, the domain expert tends to "over-label" faults (i.e., marking long segments as faulty even though only some of the points in those segments are actually faulty). We correct for this behavior when directly comparing our classification to the expert's by introducing a widening method. For any point labeled as anomalous (Bad or Very Bad) by both the expert and our system, we widen our classification by assigning the same class to $\lambda$ points before and after the current quarter hour. We apply this widening only to those anomalous types we consider to be non-trivial as the expert is very precise in labeling of trivial anomalies.

## 5.4  PRECISION AND RECALL

Figure 8 displays the precision and recall results for each of the meteorological stations over a range of $\lambda$ values for the aforementioned widening method. The diamond-marked line represents the average performance among all three sites. It is clear that the

Central met. station benefits the most from widening, which is a result of that site containing many medium-type anomalies (broken sun-shield predominately). This method allows us to tune our system to catch more anomalies at the cost of some precision. For example, by increasing $\lambda$ from 0 (individual labeling) to 200 (2 day window on either side of the current quarter hour), we have increased our average recall from 55% to 77%; however, we have done so at the cost of increasing our FP-rate from 2.4% to 5.3%.

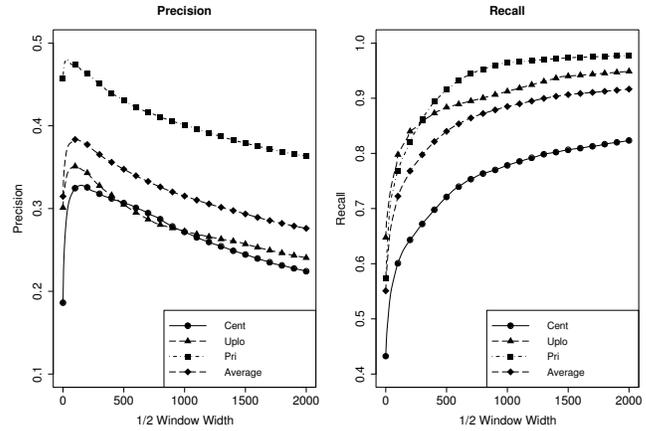

Figure 8: Precision and Recall as a function of $\lambda$. Marked increments of 100.

Comparing our model results to the domain expert labels across all data sets, we obtain the following: Precision = 0.377, Recall = 0.768, and False-Positive Rate (Specificity) = 0.053. However, these overall numbers hide many different situations. The plot in Figure 9 shows the precision and recall values for all 48 test sets (four years of data, 12 data streams per year) evaluated at $\lambda = 200$. The boxes in the figure identify interesting clusters of behavior. The largest cluster of situations is indicated by box (a). These 17 data sets have a recall > .9 and precision < .2. They represent years having 0 to 314 anomalies (median= 13). Most of these anomalies are simple and infrequent. Given our FP-Rate (on average $.053 \times 35040 = 1857$ FPs per year), the abundance of such data sets diminishes our overall precision. However, as we are maintaining near-100% recall for these years, we are still reducing by over 94% the amount data that the domain expert must review manually without missing any anomalies.

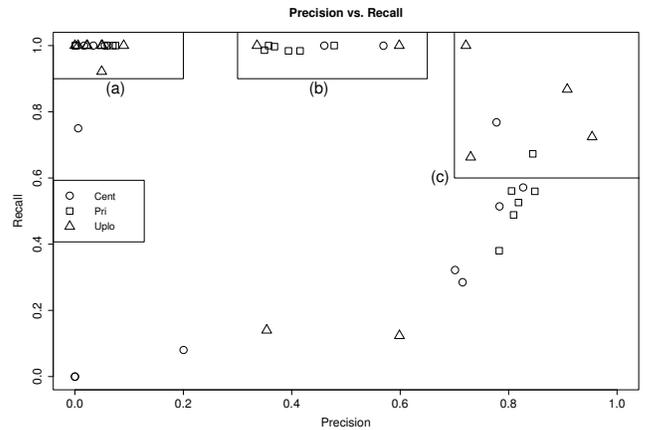

Figure 9: (a) infrequent, simple anomalies, (b) short-duration simple and medium anomalies, and (c) long-duration simple and medium anomalies.

The 11 data sets indicated by box (b) all contain short continuous periods of anomalies caused by voltage errors (in the case of the Primary Met errors) or sunshield/snow-dampening medium anomaly types (Central Met and Upper Lookout). The median anomaly count (TP + FN) in data sets in (b) is 628, which increases precision considerably compared to (a) by mitigating the effect of the FP-rate.

Box (c) contains 6 data sets with recall > .6 and precision > .7. These data sets contain long-duration simple and medium anomalies (median= 11116). The 2 years from Upper Lookout with precision > .8 denote years in which the 1.5m sensor was disconnected for the winter, resulting in a stream of faulty values. The 2 Upper Lookout years with precision ≈ .7 denote medium anomaly types caused by snow-pack around the 1.5m sensor and a non-trivial voltage range error.

The points immediately below (c) in the scatterplot show a precision-recall trade-off associated with sensor-swapped values. Specifically, years with prolonged sensor swaps tend to yield higher precision scores due to most of the values being labeled anomalous by the domain expert. As a result, any point our system labels as anomalous will be matched by the expert labeling; however, recall remains low for these years, as our system is as of yet unable to detect all of these complex anomalies. The two years having 0 precision/recall each contain a single anomalous value that our system failed to detect and have yet to be explained by the domain expert.

## 6   CONCLUDING REMARKS

The domain expert is very pleased with the performance of the model. Virtually all existing data QA tools only work by comparing multiple data streams. The fact that our model can find medium-difficulty anomalies (such as sun shield failures) by analyzing only a single data stream is surprising. We are currently deploying the model at the H. J. Andrews LTER site. The raw data will be processed by the model and then immediately posted on the web site (along with a disclaimer that an experimental automated QA process is being used). This will significantly enhance the timeliness and availability of the data. The manual QA process will still be performed later, but using the model to focus the expert's time and attention.

To detect more subtle anomaly types (swapping between sensor leads, snow pack, etc.), our model must exploit the spatiotemporal correlations between sensors. In addition to providing a means of detecting additional sensor faults, this extension to the model will allow us predict with greater certainty the actual temperature in situations where at least two of the four sensors are functioning correctly.

## Acknowledgments

We would like to thank Frederick Bierlmaier and Donald Henshaw for providing us with the raw and processed atmospheric data from the H.J. Andrews LTER and for their help in discerning the anomaly types found therein. This work was supported under the NSF Ecosystem Informatics IGERT (grant number DGE-0333257).